\pdfoutput=1

\documentclass[11pt]{article}

\usepackage[preprint]{acl}

\usepackage{times}
\usepackage{latexsym}
\usepackage{graphicx}
\usepackage[normalem]{ulem}
\usepackage{booktabs}
\usepackage{arydshln}
\usepackage{multirow}
\usepackage{lscape}
\usepackage[most]{tcolorbox}

\usepackage{amsmath}
\DeclareMathOperator*{\argmax}{arg\,max}

\usepackage[T1]{fontenc}

\usepackage[utf8]{inputenc}

\usepackage{microtype}

\usepackage{inconsolata}

\definecolor{km}{HTML}{0000FF}

\title{Getting Serious about Humor: \\ Crafting Humor Datasets with Unfunny Large Language Models}

\author{Zachary Horvitz\textsuperscript{\rm 1,*}, Jingru Chen\textsuperscript{\rm 1,*}, Rahul Aditya\textsuperscript{\rm 1}, Harshvardhan Srivastava\textsuperscript{\rm 1},  \\ \textbf{Robert West\textsuperscript{\rm 2}}, \textbf{Zhou Yu\textsuperscript{\rm 1}}, \textbf{Kathleen McKeown\textsuperscript{\rm 1}} \\ \textsuperscript{\rm 1}Columbia University, \textsuperscript{\rm 2}EPFL
\\ \\
\texttt{\{zfh2000, jc5898, ra3261, hs3447, zy2461\}@columbia.edu} \\ \texttt{robert.west@epfl.ch, kathy@cs.columbia.edu}
  }

\begin{document}
\maketitle

\def\thefootnote{*}\footnotetext{Equal contribution.}\def\thefootnote{\arabic{footnote}}

\begin{abstract}
Humor is a fundamental facet of human cognition and interaction. Yet, despite recent advances in natural language processing, humor detection remains a challenging task
that is complicated by the scarcity of datasets that pair humorous texts with similar non-humorous counterparts. We investigate whether large language models (LLMs) can generate synthetic data for humor detection via editing texts.
We benchmark LLMs on an existing human dataset and show that 
current LLMs display an impressive ability to ``unfun'' jokes, as judged by humans and as measured on the downstream task of humor detection. We extend our approach to a code-mixed English-Hindi humor dataset where we find that \textsc{Gpt-4}'s synthetic data is highly rated by bilingual annotators and provides challenging adversarial examples for humor classifiers.

\end{abstract}

\section{Introduction}

Despite their success on natural language tasks, large language models (LLMs) struggle to reliably detect and explain humor \cite{baranov-etal-2023-told, GesIsGG, hessel-etal-2023-androids}, and generate novel jokes \cite{jentzsch2023chatgpt}.  Notably, humans also struggle to write jokes; even at satirical newspapers like \textit{The Onion}, less than $3\%$ of proposed headlines are printed \cite{west-horvitz-aaai2019-unfun, Glass_2008}. In contrast, humans are able to consistently edit jokes to \textit{unfun} them, an insight which motivated \citet{west-horvitz-aaai2019-unfun} to host a game where internet users competed to edit satirical headlines to make them serious. 
The resulting dataset, the \textit{Unfun Corpus} \cite{west-horvitz-aaai2019-unfun},  has been a valuable tool for advancing computational humor research. The dataset has been used to study properties of both humor and transformer architectures \cite{west-horvitz-aaai2019-unfun, peyrard2021laughing} and even to generate novel satire \cite{horvitz-etal-2020-context}. Additionally, recent work has found that despite the relatively small size of the original dataset, humor detection models trained on Unfun data generalize remarkably well to other datasets,
while models trained on other humor datasets perform poorly at classifying Unfun-edited data \cite{baranov-etal-2023-told}. 

\begin{figure}[t]
\includegraphics[width=7.5cm]{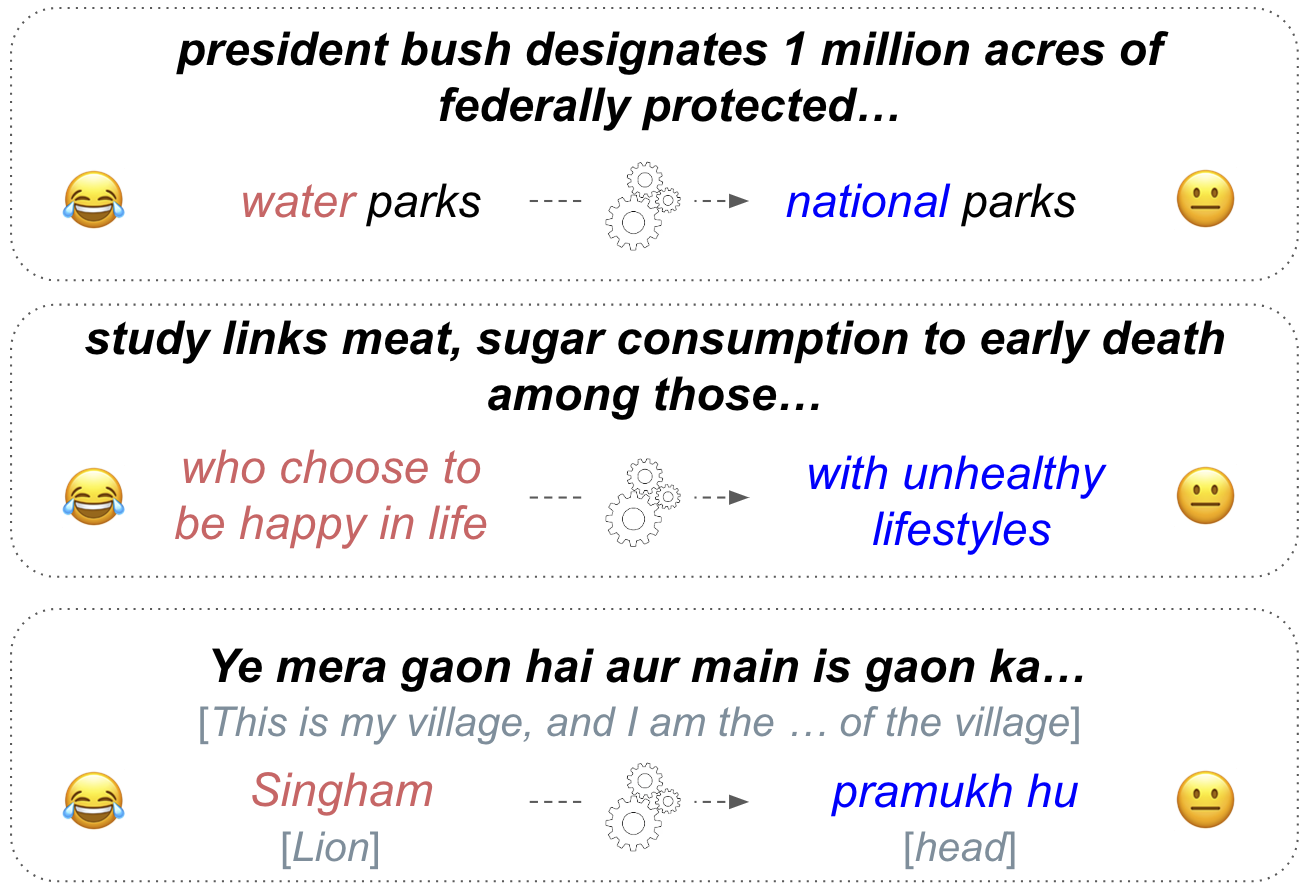}
\centering
\caption{Outputs from \textsc{Gpt-4}. We leverage language models to \textit{edit away} (or "unfun") humor in existing human-written jokes, resulting in aligned datasets that pair humorous texts with non-humorous counterparts. }
\label{fig:figure1}
\end{figure}

While useful contributions, Unfun and other aligned humor datasets \cite{hossain-etal-2019-president, hossain2020stimulating} are limited in both size and scope, due to their reliance on human annotation. 
We investigate the alternative of using LLMs to create datasets of aligned humorous and non-humorous texts.\footnote{Our code and datasets will be made available at \url{https://github.com/zacharyhorvitz/Getting-Serious-With-LLMs}. }  Previous work \cite{jentzsch2023chatgpt, Li2023SyntheticDG, Veselovsky2023GeneratingFS} has found that LLMs are limited in their ability to create synthetic humor. We take a new approach, exploiting the asymmetrical difficulty
\cite{josifoski-etal-2023-exploiting} of synthetic humor generation. Rather than only testing whether LLMs can \textit{generate} humor, we explore their ability to \textit{edit away} humor in existing jokes.  Validating and harnessing this capability could provide large paired datasets and support future work on improving humor detection and even generation.

Our contributions include benchmarking against human-curated data in the Unfun corpus, where we find that LLMs like \textsc{Gpt-4} and \textsc{Gpt-3.5} \cite{openai2023gpt4,chatgpt} can (1) outperform humans at removing humor from texts and that (2) this ability can be harnessed to generate high quality synthetic data for training humor classifiers. While these models \textit{can} also be prompted to modify unfunny headlines to craft satire, we find that this ability is more inconsistent and lags behind satirical writers. Finally, we consider a code-mixed English-Hindi humor dataset to evaluate whether \textsc{Gpt-4}'s ``unfunning'' ability generalizes to other domains and languages. We find that the resulting synthetic unfunny dataset is rated highly by bilingual annotators and poses challenging adversarial data for models trained on the original corpus.

\section{Getting Serious with Language Models}

We first revisit the Unfun task and resulting dataset, but with language models as players. 

\subsection{Unfun Dataset}
In the original Unfun game \cite{west-horvitz-aaai2019-unfun}, players were tasked with editing existing satirical headlines from \textit{The Onion},\footnote{\url{https://www.theonion.com/}} to transform the original satire into corresponding serious headlines. For example (removing ``Delicious''):
\begin{center}
    ``Scientists Discover \sout{Delicious} New Species"
\end{center}
Players were rewarded for preserving token-level similarity with the original satire and for crafting convincingly  serious headlines that other players rated as real. The resulting dataset includes approximately 11K unfunned headlines, with a subset rated by players.  We leverage Unfun pairs, of satirical headlines and their unfunned counterparts, to benchmark the performance of LLMs at editing humorous texts against humans. We include additional details on data preparation in Appendix \ref{sec:unfun dataprep}.

\subsection{Unfun Generation}

We consider a few-shot setting \cite{brown2020language}, and provide LLMs with a short task description, along with a set of input-output exemplar pairs: \textit{(humorous text, serious text)}. Following \citet{Veselovsky2023GeneratingFS}, we encourage diversity in our synthetic data by sampling these exemplars from a subset of the existing pairs rated as high-quality by the original human players.
For the unfunning task, we consider four popular LLMs: \textsc{Gpt-4} \cite{openai2023gpt4} and \textsc{Gpt-3.5-turbo}, along with \textsc{Mistral-7b-instruct} and \textsc{Mistral-7b} \cite{jiang2023mistral}. 
\begin{table*}[t]
\centering
\begin{tabular}{@{}clcccc@{}}
\toprule
& & \multicolumn{2}{c}{Data Characteristics} & \multicolumn{2}{c}{Holdout Accuracy} \\ 
\cmidrule(lr){3-4} \cmidrule(lr){5-6}
Direction & Source & Diversity (TTR) & Edit Dist & \textsc{Mistral} & \textsc{RoBERTa} \\ 
\midrule
\multirow{7}{*}{\textbf{Unfun}}
& \textsc{roberta-swap} & 0.262 & 2.7 & 69.9 (0.9) & 62.7 (0.7) \\
& \textsc{Mistral} & 0.257 & \textbf{2.1} & 70.7 (0.7) & 61.7 (0.3) \\
& \textsc{Mistral Instruct} & 0.255 & \underline{2.4} & 70.9 (0.7) & 64.7 (0.5) \\
& \textsc{Gpt-3.5} & 0.259 & 4.5 & 72.9 (0.2) & 65.9 (0.4) \\
& \textsc{Gpt-4} & 0.252 & 3.8 & \underline{76.5} (0.2) & \underline{69.9} (0.5) \\
\cline{2-6}
& News Headlines & \textbf{0.306} & - & 66.3 (0.2) & 64.1 (0.2) \\
& Unfun Players & \underline{0.271} & 2.9 & \textbf{80.3} (0.5) & \textbf{72.7} (0.4) \\
\midrule
\multirow{6}{*}{\textbf{Humor}}
& \textsc{Mistral} & 0.244 & 2.8 & 66.3 (0.7) & 56.3 (0.4) \\
& \textsc{Mistral Instruct}  & 0.221 & 4.5 & 65.2 (0.8) & 58.8 (0.4) \\
& \textsc{Gpt-3.5} & 0.24 & 4.6 & 69.9 (0.5) & 58.7 (0.4) \\
& \textsc{Gpt-4} & 0.246 & 5.5 & 69.5 (0.7) & 59.7 (0.6) \\
\cline{2-6}
& The Onion & 0.262 & - & - & - \\
\bottomrule

\end{tabular}
\caption{Automatic evaluations of synthetic Unfun data. We consider the two directions of editing away (\textbf{Unfun}) and editing in humor (\textbf{Humor}). We report median accuracies (and standard error) on a balanced holdout set ($n=750$) over $5$ seeds when fine-tuning \textsc{Mistral} \cite{jiang2023mistral} and \textsc{RoBERTa} \cite{liu2019roberta} humor classifiers.
} 
\label{table1}
\end{table*}
\begin{table*}[t]
\centering
\begin{tabular}{@{}llccccc@{}}
\toprule
Direction & Source & Rated Real &  \textit{Slightly} Funny /  Funny & Grammatical & Coherence \\ 
\midrule
\multirow{6}{*}{\textbf{Unfun}}
& \textsc{roberta-swap} & 30\% & \underline{15\%} /  5\% & 93\% & 86\% \\
& \textsc{Mistral Instruct} & 21\% & 50\% / 14\% &  \textbf{100\%} & 96\% \\
& \textsc{Gpt-3.5} & \underline{51\%} & 23\% /  \underline{3\%} & \textbf{100\%} & 98\% \\
& \textsc{Gpt-4} & 49\% & 21\% /  \underline{3\%} & \textbf{100\%} & \textbf{99\%} \\
\cline{2-7}
& News Headlines & \textbf{81\%} & \textbf{2\%} /  \textbf{0\%} & 99\% & 93\% \\
& Human Players & 33\% & 21\% /  7\% & 94\% & 92\% \\
\midrule
\multirow{5}{*}{\textbf{Humor}}
& \textsc{Mistral Instruct} & 21\% & 34\% /  9\% & 99\% & 93\% \\
& \textsc{Gpt-3.5} & 11\% & \underline{54\%} /  8\% & \textbf{100\%} & 94\% \\
& \textsc{Gpt-4} & \underline{10\%} & 45\% /  \underline{10\%} & \textbf{100\%} & \textbf{98\%} \\
\cline{2-7}
& The Onion & \textbf{4\%} & \textbf{68\%} /  \textbf{24\%} & 99\% & \underline{97\%} \\
\bottomrule
\end{tabular}
\caption{Human evaluations of synthetic Unfun data. We consider $n=100$ samples per approach. We collect three annotations per example and assign labels by majority agreement.} 
\label{table2}
\end{table*}

We also consider a lightweight alternative approach, \textsc{roberta-swap}, that replaces low probability tokens using predictions from a \textsc{RoBERTa} masked language model \cite{liu2019roberta}. This approach is motivated by the Incongruity Theory of Humor \cite{hutcheson1750reflections, sep-humor}, which associates humor with surprise, and previous work that has found humorous headlines to have higher perplexities \cite{peyrard2021laughing}. \textsc{roberta-swap} edits satirical headlines by iteratively performing token swaps at $k$ positions. At each selected position, the original token is replaced with the highest probability token predicted by the model at that masked time-step. The $k$ swap positions are selected using the ratio between the probability of the original token and the probability assigned to the language model's prediction. Additional details on unfun generation are included in Appendix \ref{sec: unfun gen}.

\section{Unfun Evaluation}

\subsection{Experimental Setup}

The existing Unfun data enables comparison of human and LLM players, via both \textbf{automatic} and \textbf{human} evaluations. We first evaluate the quality of synthetically generated data through automated evaluation on the downstream task of Unfun detection, and then follow this with a human evaluation.

\subsubsection{Automatic Evaluations}

First, following recent work on synthetic data \cite{Li2023SyntheticDG, Veselovsky2023GeneratingFS} we evaluate the data quality of outputs from LLMs by testing whether binary humor classifiers trained
on the synthetic outputs can differentiate between actual humorous and unfunned headlines from the original Unfun dataset.  We compare training on data from human players and actual satirical headlines
to two configurations of synthetic data: 
\begin{center}
    [\textit{Synthetic} unfun; Original satire]
    
    [Human unfun; \textit{Synthetic} satire]
\end{center}
These two configurations enable comparing the ``unfunning'' and joke writing capabilities of LLMs. Additionally, we consider the alternative of using actual unrelated news headlines as non-humorous examples. 
Using data from each approach, we fine-tune \textsc{RoBERTa} and \textsc{Mistral-7b} for humor classification.
Our test set comprises a subset of headline pairs from the Unfun corpus that were highly rated in the original game. Additional evaluation details are provided in Appendix \ref{sec:autoeval}.
 
\subsubsection{Human evaluations}

To perform our human evaluations, we recruited $10$ university students as annotators, all of whom were American and native English speakers.
Annotators were tasked with rating headlines as \textit{real/satire/neither}. In the case of the ``satire'' label, we also task the annotators with rating \textit{funniness} ($[0=\textit{not funny}, 1=\textit{slightly humorous}, 2=\textit{funny}]$). If the annotator selects ``neither'', we ask them to rate the headline's \textit{grammaticality} ($\{0,1\}$) and \textit{coherence} ($\{0,1\}$). We gather three annotations for each sample and assign labels based on majority vote.
We include additional information on our human evaluations and annotation scheme in Appendix \ref{sec: humaneval} and \ref{sec:unfunAnnotationInstructions}

\subsection{Results}
\textbf{Automatic Evaluations}
Table \ref{table1} contains the automatic evaluations on the Unfun corpus. Notably, when validated on human data, humor classifiers trained on \textsc{Gpt-4}'s synthetic unfun data are very performant, incurring the smallest accuracy drop relative to human-edited training data ($\Delta_{\textit{Mistral}}=-3.8\%$ and $\Delta_{\textit{RoBERTa}}=-2.8\%$). In contrast, classifiers trained with real news headlines as unfunny data perform poorly, highlighting the importance of aligned data for this task.  However, we find that not all aligned data is created equal, and that classifiers perform significantly worse when trained on synthetic \textit{humor} data relative to human-edited data ($\Delta<-10\%$). Even data from our \textsc{roberta-swap} unfun baseline dramatically outperforms, or is on par with, all synthetic humor approaches. The edit distances demonstrate that each approach retains a large portion of the original humorous text. However, \textsc{Gpt-4} and \textsc{Gpt-3.5} tend to modify headlines more than human players ($3.8$ and $4.5$ vs $2.9$).

\noindent \textbf{Human Evaluations} Table \ref{table2} displays the results from our human evaluations. All approaches for generating synthetic humor significantly underperform \textit{Onion} headlines on funniness and realness ratings $(p < 0.05)$. Notably, we do not observe a significant improvement between \textsc{Gpt-3.5} and \textsc{Gpt-4}. In contrast, synthetic unfuns from both \textsc{Gpt-3.5} and \textsc{Gpt-4} were significantly more likely than human unfuns to be rated as real news headlines. They were also rated as similarly unfunny and more grammatical and coherent. Surprisingly, our simple \textsc{roberta-swap} approach also performed comparably with Unfun players on funniness and real headline metrics, but underperformed on coherence.  Together, these results indicate that current LM-based methods underperform satirical writers on \textit{humor generation}, but can outperform human crowd-workers at \textit{editing away} humor in satire to craft aligned datasets.

\section{Extending Unfun to Other Languages}

\begin{table}
\setlength{\tabcolsep}{4pt} 
\centering
\begin{tabular}{lccc}
\toprule
 Source & Edit Dist & Humor & Coherence \\ 
\midrule
Non-Humor & - & 16.8\% & 92.8\% \\
\textsc{Gpt-4} Unfuns & 6.6 & 16.0\% & 93.6\%  \\
+ \textsc{Gpt-4} Filter & 6.9 & 3.6\% & 89.3\% \\

\cline{1-4}
Humor & - & 48.0\% & 93.6\% \\

\bottomrule
\end{tabular}
\caption{Human evaluations and edit distance of original and synthetic English-Hindi Tweet data \cite{khandelwal2018humor}. $n=125$ per approach.   }
\label{table3}
\end{table}

\begin{table*}
\centering
\begin{tabular}{@{}lcccc@{}}
\toprule
& & \multicolumn{3}{c}{Original Dataset} \\
\cmidrule(lr){3-5}
Source & Unfuns & Balanced Accuracy & Humor & Non-Humor \\
\midrule
Original & 22.6 (3.7) & 67.9 (0.9) & 80.3 (3.5) & 56.9 (5.1) \\
(25\%) Synth Unfuns & 34.0 (8.4) & 67.7 (1.7) & 78.4 (3.3) & 55.4 (5.9) \\
(50\%) Synth Unfuns & 57.7 (6.0) & 62.1 (0.6) & 68.4 (5.7) & 55.9 (4.7) \\
\bottomrule
\end{tabular}
\caption{Automatic evaluations with English-Hindi synthetic data. We report median accuracies (and standard error) on a holdout set from the original dataset ($n=591$) and the human-vetted unfuns ($n=97$). We also report median class-level accuracies for the original dataset.}  
\label{table4}
\end{table*}

Recent work has found that \textsc{Gpt-4} exhibits strong multilingual capabilities \cite{Mller2023IsAP, Jiao2023IsCA, Ahuja2023MEGAME}. Motivated by these findings, we investigate whether its ability to edit away humor generalizes to other languages and forms of joke.

\subsection{Experimental Setup}

We consider an existing corpus of code-mixed English-Hindi tweets, previously annotated as humorous or non-humorous \cite{khandelwal2018humor}. Here, we prompt \textsc{Gpt-4} to unfun humorous tweets. To remove low quality results, we secondarily filter outputs that \textsc{Gpt-4} still classifies as humorous. We provide additional details on dataset preparation in Appendix \ref{sec:english-hindi dataset} and English-Hindi unfun generation in \ref{sec:datagen}.

We perform a \textbf{human evaluation} with bilingual annotators who rated these unfunned outputs from \textsc{Gpt-4} alongside samples from the original dataset. We also run an \textbf{automatic evaluation}, testing the performance of humor classifiers trained with different proportions of synthetic non-humorous data. We evaluate on holdout synthetic data rated by the annotators as coherent and successfully non-humorous. For the humor classifier, we fine-tune an \textsc{xlm-roberta} model \cite{conneau2020unsupervised} previously fine-tuned on English-Hindi Twitter data \cite{nayak-joshi-2022-l3cube}.

\subsection{Results}

Tables \ref{table3} and \ref{table4} contain the human evaluations and automatic results for English-Hindi data. \textsc{Gpt-4} edited texts were rated comparably to non-humorous human tweets 
despite being derived from humorous tweets, which were rated as humorous by our annotators ($48\%$) of the time. Filtering with \textsc{Gpt-4} yielded a smaller sample ($56$/$125$) that was rated as much less humorous ($3.6\%$). These results demonstrate that \textsc{Gpt-4} is able to reliably unfun English-Hindi tweets, but with more edits than American satirical headlines ($6.6$ vs $3.8$).
 Additionally, unfunned data can provide a challenging adversarial dataset. In Table \ref{table4} we evaluate the performance of humor classifiers on human-vetted unfunned data. When trained on the original dataset, the classifier fails to generalize to the unfunned samples and performs poorly ($23\%$ accuracy).
 Incorporating synthetic training data improves this metric at a cost to accuracy on humorous examples in the original dataset. Together, these results provide evidence that the humor classifier relies on superficial features to identify humorous text, and that, even with fine-tuning, the model struggles to recognize synthetic unfunny data.

\vspace{18pt}

\section{Discussion}

Our results indicate that current LLMs struggle to generate humor, but can outperform crowd-workers at editing away (or \textit{unfunning}) humor.  We hypothesize that maximum likelihood training, combined with autoregressive sampling techniques, does not endow models with the creative spark required for joke writing, and instead lends itself to making high probability, reasonable substitutions to replace incongruous twists.
Our evaluations on code-mixed English Hindi Twitter data indicate that, for \textsc{Gpt-4}, this ability can impressively generalize to other languages and settings to create novel Unfun-like datasets. We are excited for future work that harnesses this capability and resulting data to improve humor detection and generation systems, and also to demystify fundamental properties of humor.

\section{Limitations}

We consider two settings, English satirical headlines and code-mixed English-Hindi tweets.
Humor practices and references vary by culture \cite{Alden1993IdentifyingGA, Jiang2019CulturalDI}, and we leave investigating cultural impacts on LLMs and humor to future work.
In both of our evaluations, the subjectivity of humor presents a challenge for our evaluations \cite{subjective}. We see evidence of this in Table \ref{table3}, where only $48\%$ of tweets previously annotated as humorous were also rated as humorous by our annotators, and where $16\%$ of non-humorous tweets were rated as humorous. This likely reflects differences in background knowledge and context between annotators. Additionally, we note that human Unfun players were incentivized to perform minimal edits, which may have affected their human evaluation metrics and lowered edit distances. On average, however, \textsc{Gpt-4} performs less than one additional word edit, and several approaches, including \textsc{roberta-swap}, were performant with lower edit distances than human players. Another concern is data contamination \cite{sainz-etal-2023-nlp}, and that a portion of the text from the Unfun corpus could have been trained on and memorized by the LLMs we evaluated. We investigate this concern in Appendix \ref{sec:memorization}. We note that our results on English-Hindi data show that \textsc{Gpt-4}'s abilities generalize to a dataset where these pairs do not already exist on the internet.

\section{Ethical Statement}

Humor brings joy to people and plays a critical role in building and maintaining social relationships \cite{basso1979portraits}. However, its importance presents a double-edged sword; offensive and hurtful humor can cause real harms, and reinforce prejudice \cite{harm}. As a result, with their widespread adoption, it will be paramount for AI systems to be more capable of identifying and appropriately navigating jokes. We believe that our work on benchmarking LLM humor abilities and building challenging detection datasets is an important step in this direction. However, one possible concern is that malicious actors could leverage our \textit{unfunning} approach to circumvent existing safeguards. In our experimentation, we found numerous settings where \textsc{Gpt-4} refused to generate jokes for offensive topics, but had no trouble editing texts to remove humor and offensiveness. This could enable building large parallel datasets of (offensive-text, non-offensive counterparts) that could then be used to train models for offensive joke generation. 

\section*{Acknowledgements}

We would like to thank Eric Horvitz for guidance that helped shape the direction of this work. We are also grateful to Nicholas Deas, Debasmita Bhattacharya, and Maximillian Chen for their feedback. Additionally, we would like to extend our gratitude to Amith Ananthram,  Samir Gadre, Fei-Tzin Lee, Matthew Toles, Elsbeth Turcan, Melanie Subbiah, Emily Allaway, Tymon Nieduzak, Rattandeep Singh, Prabhpreet Singh Sodhi, and Apoorva Joshi for support on human evaluations.

\bibliography{custom}

\appendix

\section{Appendix}
\label{sec:appendix}

\subsection{Data Preparation}
\label{sec:dataprep}
\subsubsection{Unfun Corpus}
\label{sec:unfun dataprep}

We use the February 2, 2023 Unfun \cite{west-horvitz-aaai2019-unfun} database backup,\footnote{\url{https://github.com/epfl-dlab/unfun}} and consider all valid unfunned headlines (i.e. not \textit{None}). This results in $11831$ pairs. A subset of these have ratings from other players. We use these to curate a \textbf{high quality} evaluation subset of pairs where:
\begin{itemize}
    \item There is at least one annotation.
    \item The satirical headline has a funniness rating $\ge 0.8$.
    \item The unfunned headline has a funniness rating $\leq 0.2$.
\end{itemize}
The resulting $867$ pairs were split among prompt examples ($10\%$), dev ($30\%$), and test ($60\%$) shards.
For our training set, we consider the remaining headlines, again ensuring that there is no overlap with other shards. The resulting dataset has many instances where there are multiple unfunned counterparts for each satirical headline. As an additional step, we randomly filter our training, dev, and test shards so that there is only one unfunned headline per satirical headline. This results in a training set of $3882$ unfuns, a dev set of $186$ unfuns, and a test set of $375$ unfuns, in each case, these are included alongside their corresponding satirical headlines. For an additional training data baseline, we also retrieve an equal number of real news headlines included in the Unfun database.

\subsubsection{Code-Mixed English-Hindi Humor}
\label{sec:english-hindi dataset}

We use the version of the English-Hindi Humor dataset
by \citet{khandelwal2018humor} hosted on GitHub.\footnote{\url{https://github.com/Ankh2295/humor-detection-corpus}}
We use the provided labels for the available data. Notably, a portion of annotated samples appear to be unavailable. We divide the available dataset ($n=2951$) into training, dev, and test shards ($60\%$, $20\%$, $20\%$). Additionally, we filter tweets containing links. 

\subsection{Data Generation Details}
\label{sec:datagen}

We include our full prompts in Appendix \ref{sec:prompts}. For decoding hyperparameters, we use $\textit{top-p}=0.85$ and $\tau=1.0$ for all LLMs. 

\subsubsection{Unfun Data Generation}
\label{sec: unfun gen}

To generate synthetic Unfun for each LLM approach, we prompt each model with $8$ randomly sampled in-context pairs from examples from our high quality subset that was set aside for prompting. For our \textsc{roberta-swap} baseline, we replace tokens in the original satirical headline using a \textsc{roberta-base}\footnote{\url{https://huggingface.co/FacebookAI/roberta-base}} model. To select each replacement, we iterate over and individually mask each token in the headline, and then predict the masked token:
\[
\hat{x}_i = \argmax_x P(x \mid x_{\neq i}, \theta_{\text{RoBERTa}})
\]
The position with the largest ratio between the predicted token and the original token probabilities is selected as the swap position:
\[
\text{swap position} = \argmax_i [\frac
{P(\hat{x}_i \mid x_{\neq i}, \theta_{\text{RoBERTa}})}
{P(x_i \mid x_{\neq i}, \theta_{\text{RoBERTa}})}
]
\]
We then replace $x_i$ with $\hat{x}_i$, and repeat this procedure $k$ times. We set $k=3$ in our experiments.

\subsubsection{Hindi-English Data Generation}
\label{sec: hindi-english gen}

Unlike for Unfun, we do not have existing pairs of (un-humorous, humorous) English Hindi tweets. To remedy this, we first generated $50$ examples in a zero-shot setting on our training set, and then selected nine high quality results to serve as our prompt. We additionally prompt \textsc{Gpt-4} with humorous and non-humourous texts to classify the resulting unfunned tweets as humorous or non-humorous. We filter unfunned tweets if they are still classified as humorous.

\subsection{Human Evaluations}
\label{sec: humaneval}

 We recruited $10$ university students as annotators for the \textbf{Unfun task}. All annotators were American and native English speakers. For the \textbf{English-Hindi} dataset, we worked with three bilingual (Hindi and English) speakers. For both evaluations, we gathered three unique annotations per example, and assigned labels based on majority votes. Our Unfun evaluation assumes that any headline labeled as satirical or as real headline is grammatical and coherent. In contrast, we do not consider the grammatical label for English-Hindi data, due to the varied syntactic styles of tweets.

In Table \ref{table2}, headlines are only rated "Real" if a majority of annotators rated the headline as "Real" (not "Satire" or "Neither"). Headlines are rated "Slightly Funny" if a majority of annotators assigned the headline $\textit{funniness}  \geq 1$, and "Funny" with $\textit{funniness}  = 2$. Our full instructions for both human evaluations are included in Appendix \ref{sec:unfunAnnotationInstructions}. Tables \ref{kripunfun} and \ref{kriphindi} display inter-annotator agreement statistics.

\begin{table}[h]
\centering
\begin{tabular}{lcc}
\toprule
\textbf{Human Label} & \textbf{Krippendorff} \\
\midrule
Real & 0.507 \\
Funny & 0.333 \\
Very Funny & 0.214 \\
Grammar & 0.271 \\
Coherence & 0.214 \\
\bottomrule
\end{tabular}
\caption{Krippendorff's $\alpha$ results on Unfun dataset.}
\label{kripunfun}
\end{table}

\begin{table}[h]
\centering
\begin{tabular}{lcc}
\toprule
\textbf{Human Label} & \textbf{Krippendorff} \\
\midrule
Coherence & 0.206 \\
Humorous & 0.377 \\
\bottomrule
\end{tabular}
\caption{Krippendorff's $\alpha$ results on English-Hindi dataset.}
\label{kriphindi}
\end{table}

\subsection{Automatic Evaluations}
\label{sec:autoeval}
On the \textbf{Unfun dataset}, for each synthetic Unfun approach, we generate data using the corresponding original $3882$ training examples as inputs. We then evaluate classifiers trained on each dataset on the filtered high quality holdout data. To generate humor, we provide the unfunned example as input. To edit away humor, we provide the original satirical headline. We also provide in-context pairs drawn from the high quality prompt examples (See \ref{sec:unfun dataprep}). For our Real News baseline, we randomly select $3882$ real news headlines to serve as non-humorous examples. 

On the \textbf{English-Hindi dataset}, we compare training on the original dataset to training on data where $(25\%)$ and $(50\%)$ of non-humorous examples have been replaced by \textsc{Gpt-4} Filtered unfunned data. We evaluate classifiers on a holdout set from original dataset ($n=591$), and also  set of Unfuns ($n=97$), derived from humorous examples in our holdout set and rated by our annotators as both coherent and non-humorous. All results for both datasets are computed over $5$ seeds.

\subsection{Humor Classifier Training}
\label{label:humorTraining}

For the Unfun task, we fine-tune \textsc{Mistral} \cite{jiang2023mistral}\footnote{\url{https://huggingface.co/mistralai/Mistral-7B-v0.1}} and \textsc{RoBERTa} \cite{liu2019roberta}\footnote{\url{https://huggingface.co/FacebookAI/roberta-base}} models. For Hindi-English, we consider \textsc{Hing-RoBERTa} \cite{nayak-joshi-2022-l3cube}\footnote{\url{https://huggingface.co/l3cube-pune/hing-roberta}}. All models are trained with the AdamW optimizer \cite{loshchilov2019decoupled} and a constant learning rate. Due to the class imbalance in the available English-Hindi dataset ($39$\% non-humorous, $61$\% humorous), we weight the loss by the inverse proportion of class frequency.

We fine-tune our \textsc{Mistral} classifier with 4-bit quantized LoRA \cite{dettmers2023qlora} and the addition of a classification head. For all classifiers, we first perform hyperparameter tuning on the original human authored datasets.

For the \textbf{Unfun dataset} we consider:

\begin{itemize}
    \item Learning Rates $\in \{5e-5, 2.5e-5, 1.25e-5, 6.25e-6, 3.125e-6, 1.5625e-6\}$
    \item Batch Size $\in [32]$ (Due to resource constraints)
\end{itemize}

For the \textbf{English-Hindi} Dataset dataset we consider:

\begin{itemize}
    \item Learning Rates $\in \{5e-5, 2.5e-5, 1.25e-5, 6.25e-6, 3.125e-6, 1.5625e-6\}$
    \item Batch Size $\in \{256, 128, 64, 32, 16, 8\}$
\end{itemize}

After selecting the highest performing configuration, we run each experiment with $5$ seeds ($[1234, 2345, 3456, 4567, 5678]$). We include the most performant hyperparameters in Table \ref{table7}.
\begin{table*}[h]
\centering
\begin{tabular}{lccc}
\toprule
 Model & Learning Rate & Batch Size \\ 
\midrule

\textsc{Mistral} (QLoRA) & 6.25e-06 & 32 \\
\textsc{RoBERTa} & 1.25e-05 & 32 \\
\textsc{Hing-RoBERTa} & 1.5625e-06 & 8 \\

\bottomrule
\end{tabular}
\caption{The training configurations for our automatic evaluations, after hyperparameter tuning.}
\label{table7}
\end{table*}
All model trains use a single NVIDIA A100 GPU. We estimate the total compute budge to be $200$ hours.

\subsection{Considering Memorization}
\label{sec:memorization}

We investigate whether data contamination and memorization is affecting our results by testing how often synthetic unfuns or humor appear in the original Unfun corpus. We find that only a small fraction of outputs appear to match human-unfunned text or satire headlines. We include results in Table \ref{table8}.
\begin{table}[h]
\centering
\begin{tabular}{lcc}
\toprule
Model & Unfun & Satire \\
\midrule
\textsc{Gpt-3.5} & 3/200 & 0/200 \\
\textsc{Gpt-4}& 7/200 & 0/200 \\
\textsc{Mistral} & 2/200 & 1/200 \\
\textsc{Mistral Instruct}  & 2/200 & 0/200 \\
\textsc{roberta-swap}  & 0/200 & - \\
\bottomrule
\end{tabular}
\caption{The number of overlapping samples between human-curated headlines and synthetic headlines in our test examples ($n=200$).}
\label{table8}
\end{table}
Of these, the majority represent simple edits, indicating that the models may have rediscovered trivial unfuns. For example:

\begin{center}
    ``Egypt plunges into state of \sout{Middle East} \textit{crisis}"
\end{center}

\section{Prompts}
\label{sec:prompts}
\subsection{Unfun Task Prompts}

\subsubsection{Humor Generation}

    \textbf{Chat Models} 

    \begin{tcolorbox}[breakable, enhanced]
    \textit{"You are a helpful assistant that edits realistic headlines to make them humorous."}
    
    \{"role": "user", "content": <Unfunned Headline>\},
    
    \{"role": "assistant", "content": <Satire Headline>\}
    
    \end{tcolorbox}

     \noindent\textbf{Completion Models} 
     
    \begin{tcolorbox}[breakable, enhanced]
        
    \textit{"The following realistic headlines can be edited to be humorous:"}

     "<Unfunned Headline> -> <Satire Headline>"

     \end{tcolorbox}

\subsubsection{Unfun Generation}

    \textbf{Chat Models} 
       \begin{tcolorbox}[breakable, enhanced]
    \textit{"You are a helpful assistant that edits humorous headlines to make them realistic."}
    
    \{"role": "user", "content": <Satire Headline>\},
    
    \{"role": "assistant", "content": <Unfunned Headline>\},

    ...
    \end{tcolorbox}

\noindent\textbf{Completion Models} 
    
    \begin{tcolorbox}[breakable, enhanced]
        
   \textit{"The following humorous headlines can be edited to be realistic:"}

     "<Satire Headline> -> <Unfunned Headline>"

\end{tcolorbox}

\subsection{English-Hindi Task Prompts}

\subsubsection{Unfun Generation}

    \textbf{Chat Models} 
    
   \begin{tcolorbox}[breakable, enhanced]
   
    \textit{"Kya ye diye hue tweet ka humor wala part hata kar use normal bana sakti ho? Aur jitna ho sake utna punctuation use same rakhne ki koshish karna" [Can you remove the humorous part of the given tweets and make them normal? And try to keep the punctuation as much the same as possible.]}.

    \{"role": "user", "content": <Context Funny Tweet>\},
    
    \{"role": "assistant", "content": <Context Unfunned Tweet>\}

\end{tcolorbox}

\subsubsection{Unfun Filtering}

\textbf{Chat Models} 

\begin{tcolorbox}[breakable, enhanced]
\textit{"You are a pattern-following assistant used to rigorously determine whether a Hindi tweet is intended to be humorous. Given a Hindi tweet, respond only with either of Yes or No. Yes if it is humoruous and No if it is not humorous"}
    
    \{"role": "user", "content": <Context Tweet>\},
    
    \{"role": "assistant", "content": <Context Yes/No Label>\}
    \end{tcolorbox}
    
\section{Human Evaluation Instructions}

\subsection{Unfun Task Instructions}
\label{sec:unfunAnnotationInstructions}

\begin{tcolorbox}[breakable, enhanced]
    \textit{Each annotator has been assigned a series of text samples to review.  First, you are asked to evaluate whether the text sounds like a}
\begin{itemize}
    \item \textit{r) real news headline (like from a non-humorous news website)}

    \item \textit{OR s) satirical news headline (like from a humorous newspaper like TheOnion.) }

    \item \textit{OR n) neither (text that would not appear in either setting, because it is ungrammatical, or incoherent.}

\end{itemize}

\textit{If you rate a headline as n (neither), you will be further prompted to rate it as a grammatical [no=0,yes=1 (for a news headline) and coherent [no=0,yes=1]. }

\textit{If you rate a headline as s (satire), you will be prompted to subjectively rate the quality of humor:} 

\begin{itemize}
    \item \textit{0 - not funny}
    \item \textit{1 - slightly humorous / there is some identifiable joke}
    \item \textit{2 - funny}
\end{itemize}

\textbf{\textit{Content Warning:
Several headlines may contain references to upsetting content.}}

EXAMPLES:
\textbf{Satirical Headlines}

\begin{itemize}
    \item nhl not quite sure why it has a preseason
    \item america's sweetheart dumps u.s. for some douchebag
    \item apple: new iphone good
    \item cat general says war on string may be unwinnable
    \item fire chief grants fireman 3-day extension on difficult fire
\end{itemize}

\textbf{News Headlines}

\begin{itemize}
    \item the word 'doofuses' may cost ex-yahoo ceo bartz \$10 million
    \item 2 meteorites hit connecticut
    \item world outraged by north korea's latest nuke test
    \item poverty rate hits 17-year high
    \item philippines: 5 foreign terror suspects in south
\end{itemize}
\end{tcolorbox}

\subsection{English-Hindi Task Instructions}

The following task instructions specify additional information based on the original instructions provided to annotators in \cite{khandelwal2018humor}.

\begin{tcolorbox}[breakable, enhanced]
\textit{Each annotator has been assigned a series of text samples to review.  
First, you are asked to evaluate whether the text is
h) humorous n) non-humorous}

\textit{Secondarily, you will be asked to rate whether a text is coherent [no=0,yes=1]
A tweet should be marked as coherent, even if you don’t have all the required background knowledge, as long as you can reasonably understand its meaning.}

\textit{Additional info:}

\begin{itemize}

    \item \textit{Any tweets stating any facts, news or reality should be classified as non-humorous.}
    \item \textit{Tweets which consisted of any humorous anecdotes, fantasy, irony, jokes, insults should be annotated as humorous}
    \item \textit{Tweets stating any facts, dialogues or speech which did not contain amusement should be put in non-humorous class.}
    \item \textit{Tweets containing normal jokes and funny quotes should be placed in the humorous category. }
    \item \textit{Some tweets consist of poems or lines of a song but modified. If such tweets contain satire or any humoristic features, then they could be categorized as humorous otherwise not.}
\end{itemize}

\textit{\textbf{Content Warning:
Several tweets may contain references to upsetting/offensive content.}}

EXAMPLES (We give the English Translations of each in brackets but they were not presented to the annotators):

\textbf{Humorous Tweets}
\begin{itemize}
    \item Jhonka hawa ka aaj bhi chhup ke hilaata hoga na \#Samir \#HawaKaJhonka \#BeingSalmanKhan [\textit{Does the breeze still sway secretly today? \#Samir \#HawaKaJhonka \#BeingSalmanKhan})
    \item Working on a Sunday, chand rupye kamaane ke liye insaan apni khushiyon ka bhi sauda kar leta hai. [\textit{Working on a Sunday, to earn a few rupees, a person sometimes even sacrifices their happiness.}]
    \item DJ wale babu bhosdike ab to gaana baja de iska.. bol bol ke kaan se khoon nikaal diya hai isne [\textit{DJ wale babu, play the song now.. he has made our ears bleed by talking so much.}]
    \item Is Arvind Kejriwal new Che Guavara ? RT @ashutosh83B Is Rahul Gandhi new Arvind Kejariwal ? [\textit{Is Arvind Kejriwal the new Che Guevara? RT @ashutosh83B Is Rahul Gandhi the new Arvind Kejriwal?}]
    \item Sukh bhare din beete re bhaiya, Babadook aayo re [\textit{Brother, may the days filled with joy pass by. The Babadook has arrived.}]
\end{itemize}

\textbf{Non-Humorous Tweets}
\begin{itemize}
    \item Apne support wale MLAs ko farmhouse main band kar lenge. Parade karayenge. Takhta palat karenge. Akhand chutiyap. [\textit{We will lock up our supporting MLAs in the farmhouse. Parade them. Flip the throne. Absolute nonsense.}]
    \item Hrithik Roshan is using Vodafone. [\textit{Hrithik Roshan is using Vodafone.}]
    \item PLEASE STOP MAKING JOKES ON SALMAN KHAN. BHAI BOLA NAHI CHALA RAHA THA GAADI TO NAHI CHALA RAHA THA. \#BHAIROXX [\textit{Please stop making jokes on Salman Khan. Bhai was not driving the car if he said he was not driving the car. \#BHAIROXX}]
    \item Bhaari sankhya mein vote karein, aapke TL par wph hi nazar aayega [\textit{Vote in large numbers, wph will only appear in your TL.}]
\end{itemize}

\end{tcolorbox}

\section{Reference Examples}

Tables \ref{table:unfun_data}, \ref{table:satire_data}, and \ref{table:hindi_data} include reference samples for English synthetic unfun outputs, English satire outputs, and English-Hindi unfun outputs respectively.

\begin{table*}[h]
\centering
\begin{tabular}
{p{1.3cm}p{3.5cm}p{3.5cm}p{3.5cm}}
\toprule
\textbf{Original Satire} & \textbf{tom petty to play some new stuff he's been working on at super bowl} & \textbf{jaguars offensive line not sure they can open big enough hole for maurice jones drew} & \textbf{obama takes surprise caller during weekly radio address} \\
\midrule
Human & tom petty to play some new stuff he's been working on at coachella & jaguars offensive line not sure they can open stable positioning hole for maurice jones drew & obama takes caller during weekly radio address \\
\textsc{Gpt-3.5} & tom petty to perform classic hits at super bowl & jaguars offensive line not sure they can create sufficient gap for maurice jones drew & obama takes surprise caller during live radio interview \\
\textsc{Gpt-4} & tom petty to perform new material at super bowl & jaguars offensive line unsure if they can open big enough hole for maurice jones drew & obama takes unexpected caller during weekly radio address \\
\textsc{Mistral} & tom petty to play some new stuff he's been working on at superbowl & jaguars offensive line not sure they can open big enough hole for joe flacco & obama takes surprise caller during weekly radio address \\
\textsc{Mistral Instruct} & tom petty to play some new songs he's been working on at super bowl halftime show & jaguars offensive line not sure they can clear big enough opening for maurice jones drew & obama takes surprise caller during weekly interview \\
\textsc{roberta-swap} & he wants to play some new stuff he's been working on at the bowl & jaguars offensive line not sure they can dig big enough hole for maurice jakes & panama takes in migrants during weekly radio address \\
\bottomrule
\end{tabular}
\caption{Unfun model output reference samples.}
\label{table:unfun_data}
\end{table*}

\begin{table*}[h]
\centering
\begin{tabular}{p{1.3cm}p{3.5cm}p{3.5cm}p{3.5cm}}
\toprule
\textbf{Original Unfun} & \textbf{obama, congress must strike a deal on budget by march 1} & \textbf{espn sports segment reveals science behind tony romo's precision pass} & \textbf{jimmie johnson disqualified after pre-race inspection reveals car to be violating rules} \\
\midrule
Human & obama, congress must reach deal on budget by march 1, and then april 1, and then april 20, and then april 28, and then may 1 & espn sports segment reveals science behind tony romo's shittiness & jimmie johnson disqualified after pre-race inspection reveals car to be airplane \\
\textsc{Gpt-3.5} & obama, congress must strike a deal on budget by march 1 or face detention & espn sports segment reveals magic behind tony romo's precision pass & jimmie johnson disqualified after pre-race inspection reveals car to be a Transformer in disguise \\
GPT-4 & obama, congress must strike a deal on budget by march 1 or risk being grounded & espn sports segment reveals magic spell behind tony romo's precision pass & jimmie johnson disqualified after pre-race inspection reveals car to be a spaceship \\
\textsc{Mistral} & obama, congress must strike a deal on budget by march 1 & espn sports segment reveals science behind tony romo's precision penis & jimmie johnson disqualified after pre-race inspection reveals car to be black \\
\textsc{Mistral Instruct} & obama, congress must strike a deal on budget by march 1, or else there will be no more kittens & espn sports segment reveals science behind kate middleton's royal wave & jimmie johnson disqualified after pre-race inspection reveals car to be violating laws and joking about it \\
\bottomrule
\end{tabular}
\caption{Satire model output reference samples.}
\label{table:satire_data}
\end{table*}

\begin{table*}[t]
\centering
\begin{tabular}{p{6cm}p{6cm}}
\toprule
\textbf{Original Humorous Tweet} & \textbf{\textsc{Gpt-4} English-Hindi Unfuns} \\
\midrule
Ab ki baar.. \#MaaBetaFarar.. & Ab ki baar.. yeh log farar hain.. \\
Husbands should be like Vim bar, gale kam aur chale zyada. & Patidev ko samarpit aur lambe samay tak saath dena chahiye. \\
O naadan parindey ghar aaja. Parinda: naadan tera baap. & O naadan parindey ghar aaja. Parinda: Mujhe ghar aane do. \\
Neend aaja nahi to kal se tujhe KRK bulaunga & Neend aaja nahi to kal se tujhe alag naam se bulaunga \\
Bhai ab itne velle bhi nahi hai ki \#IndVsBan test match dekhenge & Bhai ab itne samay nahi hai ki \#IndVsBan test match dekhenge \\
Asli toofan andar hai, jail ke andar. \#SalmanVerdict & Asli samasya jail ke andar hai. \#SalmanVerdict \\
Vodafone use karne se acha to ek kabootar pal lo. & Vodafone use karne se acha to kisi aur network provider ka use karo.\\
\bottomrule
\end{tabular}
\caption{GPT-4 English-Hindi unfunned reference samples. See Table \ref{table:hindi_translation_data} for English translations.}

\label{table:hindi_data}
\end{table*}

\begin{table*}[t]
\centering
\begin{tabular}{p{6cm}p{6cm}}
\toprule
\textbf{Original Humorous Tweet} & \textbf{\textsc{Gpt-4} English-Hindi Unfuns} \\
\midrule
This time.. \#MotherSonGone.. & This time.. these people are gone.. \\
Husbands should be like Vim bar, less talk and more work. & Husbands should be dedicated and support for a long time. \\
Oh naive bird, come home. Bird: Your dad is naive. & Oh naive bird, come home. Bird: Let me come home. \\
If sleep doesn't come, from tomorrow I will call you KRK. & If sleep doesn't come, from tomorrow I will call you by a different name. \\
Bro, we're not that free to watch the \#IndVsBan test match. & Bro, we don't have that much time to watch the \#IndVsBan test match. \\
The real storm is inside, inside the jail. \#SalmanVerdict & The real problem is inside the jail. \#SalmanVerdict \\
It's better to raise a pigeon than to use Vodafone. & It's better to use another network provider than Vodafone.\\
\bottomrule
\end{tabular}
\caption{Translation of \textsc{Gpt-4} English-Hindi unfunned reference samples.}
\label{table:hindi_translation_data}
\end{table*}

\end{document}